\documentclass[11pt]{article}

\usepackage[a4paper,margin=1in]{geometry}
\usepackage{graphicx}
\usepackage{booktabs}
\usepackage[round]{natbib}
\usepackage{url}
\usepackage{amsmath}
\usepackage[hidelinks]{hyperref}
\usepackage[capitalize]{cleveref}
\usepackage{rotating}

\crefname{section}{Sec.}{Secs.}
\Crefname{section}{Section}{Sections}
\Crefname{table}{Table}{Tables}
\crefname{table}{Tab.}{Tabs.}
\Crefname{figure}{Figure}{Figures}
\crefname{figure}{Fig.}{Figs.}

\title{TRAS: An Interactive Software for Tracing Tree Ring Cross Sections}

\author{
Henry Marichal$^{1,2,*}$,
Diego Passarella$^{3}$,
and Gregory Randall$^{1}$\\[0.5em]
\small $^{1}$Instituto de Ingeniería Eléctrica, Facultad de Ingeniería, Universidad de la República, Montevideo, Uruguay\\
\small $^{2}$Pento.ai, Montevideo, Uruguay\\
\small $^{3}$Procesos Industriales de la Madera, CENUR Noreste, Universidad de la República, Tacuarembó, Uruguay\\
\small $^{*}$Corresponding author: \texttt{hmarichal93@gmail.com}
}

\date{}

\begin{document}

\maketitle

\begin{center}
\textbf{Author Accepted Manuscript}
\end{center}

\noindent This manuscript has been accepted for publication in
\textit{Forestry: An International Journal of Forest Research}, published by Oxford University Press.
This is an author-produced version and may differ from the final Version of Record.
The final published version will be available through the journal website.

\vspace{1em}

\begin{abstract}
Tree ring marking is a key step in dendrometry, essential for studies in dendrochronology, forest dynamics, and growth analysis. However, it is still often done manually, which makes the process time-consuming, subjective, and hard to scale to large image datasets.

Here, we present the Tree Ring Analyzer Suite (TRAS), an open-source software tool with a graphical interface that enables automatic delineation and manual correction of tree rings in wood cross-sectional images. The software incorporates three complementary algorithms for automatically detecting tree rings: a classic image processing method (CS-TRD) and two deep-learning approaches (DeepCS-TRD and INBD). TRAS provides an intuitive graphical user interface that allows users to refine automatic ring detection results, remove erroneous detections, and manually add undetected rings. The software also calculates various dendrochronological metrics, including earlywood and latewood areas, ring perimeter, and equivalent ring width, with options for custom measurements along specified paths. A dataset of 18 expertly annotated tree cross-section images of \textit{Pinus taeda L.} was used for validation and accuracy assessment, with additional manual measurements serving as ground truth. For one-dimensional measurements, ring width, a comparative analysis with a traditional tool, CooRecorder, was performed.

TRAS achieved accurate automatic detection of tree rings in \textit{P. taeda} samples, with DeepCS-TRD obtaining the highest F-score, 81.0\%, and precision, 86.4\%. Automatic detection reduces the annotation effort to approximately 20\% of ring boundaries requiring manual correction, representing a substantial time saving over fully manual delineation. TRAS measurements showed excellent agreement with CooRecorder, $r > 0.99$, and common detection errors, such as jump propagation or false positives near knots, were easily corrected using the postprocessing interface.

TRAS offers a robust and flexible solution for researchers in dendrometry, enhancing the efficiency and accuracy of tree ring analysis. The software runs on Windows, macOS, and Linux. Code is available for further improvement and use at \href{https://hmarichal93.github.io/tras}{TRAS project website}.
\end{abstract}

\maketitle

\section{Introduction}


In dendrometry studies, ring width is one of the primary parameters of interest, and it is common practice to take ring width measurements from core samples. However, in tree trunks with high pith eccentricity, ring area often correlates better with climate conditions such as precipitation or temperature than ring width \citep{atmos14020319}. Hence, methods to measure tree ring area from wood discs may provide advantages in some studies over more straightforward ring width measurements.

The measurement of both tree ring area and tree ring widths has increasingly been automated over the last few years. Advancements in image processing algorithms played an important role in this development. A common workflow is that bore cores and wood discs are being digitized with a scanner or a camera and then processed in an image processing software. Storing samples in digital format (as images) facilitates sharing them with the community and validating results. Despite mentioned advancements, the methods for digitizing tree-ring measurements are still predominantly manual. One significant challenge is the difficulty of digitizing disks or cores. High-resolution scanners (up to 3.9 microns per pixel) are available, but they require samples to be completely flat, a condition difficult to achieve with cores or disks. Digital photography has emerged as a viable alternative for tree ring studies \citep{capturering}. For example, \citep{Kim_2023} compares measurements taken from scanned and smartphone images, obtaining comparable results. However, proper camera calibration is essential when using images acquired with a camera.  

Ring curves can be manually delineated from images using tools with graphical user interfaces, such as ImageJ or Adobe Illustrator \citep{Constantz_2021}, which allow users to measure areas defined by their input. Moreover, \cite{wiad} developed the Wood Image Analysis and Dataset (WIAD) tool, an R package with a graphical user interface that allows manual ring-width measuring along a linear path and archive functionalities to store the image, the ring-width series, and metadata.
However, manually delineating the complete rings in the stem using this approach is time-consuming.

\cite{Constantz_2021} proposed a Python package to study growth ring patterns using 2D information in species with asymmetric growth characteristics, such as Hawaiian Sandalwood (\textit{Santalum paniculatum}). The Python package  automatically computes multiple transects based on manual tree-ring annotations. However, a limitation of their approach is the lack of an automatic or semi-automatic method for ring delineation, making it more time-consuming than measuring ring width on core samples. 

Some graphical software tools implement automatic ring detection based on edge information in cores rather than cross-section images \citep{Kim_2023, Shi_2019, Maxwell_2021}. These tools allow users to correct detection errors and perform tree-ring measurements.

Recent developments in core sample image analysis include deep learning techniques where semantic segmentation networks try to segment the pixels that belong to the tree rings. \cite{Fabija_ska_2018} reported a recall of $96\%$ and a precision of $97\%$ on scanned images of ring-porous wood core samples using a U-Net architecture \citep{unet}, but they did not make the code available. \cite{Pol_ek_2023} trained a Mask R-CNN network over microscopy images of Norway Spruce (\textit{Picea abies L.}), reporting a recall of $98\%$ and a precision of $96\%$ over the test set (1909 ring boundaries). Code is available, and detections can be exported in the \textit{pos} format to be edited by the CooRecorder tool \citep{Maxwell_2021}. \cite{WU2024552} applied a Feature Pyramid Network (FPN) with the encoder of ResNet-18 for tree ring segmentation in hardwood species, reporting an F-Score of $73\%$. \cite{unettransformer} applied a UNETR network to segment tree rings in microscopy images, reporting that their model performed equal or better than manual tree ring boundary delineation on stained wood microsections in $92\%$ of the tested cases.

Regarding cross-sectional images, \cite{inbd} addressed the challenge of detecting tree rings in microscopy cross-section images of shrub branches  by introducing the Iterative Next Boundary Detection (INBD) method, which employs a U-Net-based architecture. In this study, the INBD method was assessed in the species \textit{Dryas octopetala L.}, \textit{Empetrum nigrum subsp. hermaphroditum} \textit{(Hagerup) Böcher}, and \textit{Vaccinium myrtillus L.}. While the source code is publicly available, the method lacks a graphical user interface for editing or correcting ring detections.

\cite{CSTRD} proposed a classical image processing approach for detecting tree rings in images of stem cross-sections of coniferous trees , named Cross Section Tree Ring Detector (CS-TRD), which was tested in \textit{Pinus taeda L.} and \textit{Picea glauca (Moench) Voss} species. This method leverages advanced edge detection techniques and includes an online demo to test the process on user-provided images. Additionally, ring curves can be exported in JSON format to be corrected with the LabelMe Tool \citep{wada2024labelme}.

Building upon CS-TRD, \cite{deepcstrd} introduced DeepCS-TRD, which replaces the edge detection step with a U-Net architecture. This enhancement allows for improved generalization across different species and image acquisition methods, outperforming CS-TRD and INBD in detecting tree rings in cross-sectional images of \textit{P. taeda} and \textit{Gleditsia triacanthos L.} species.

\begin{figure}
    \centering
    \includegraphics[width=\linewidth]{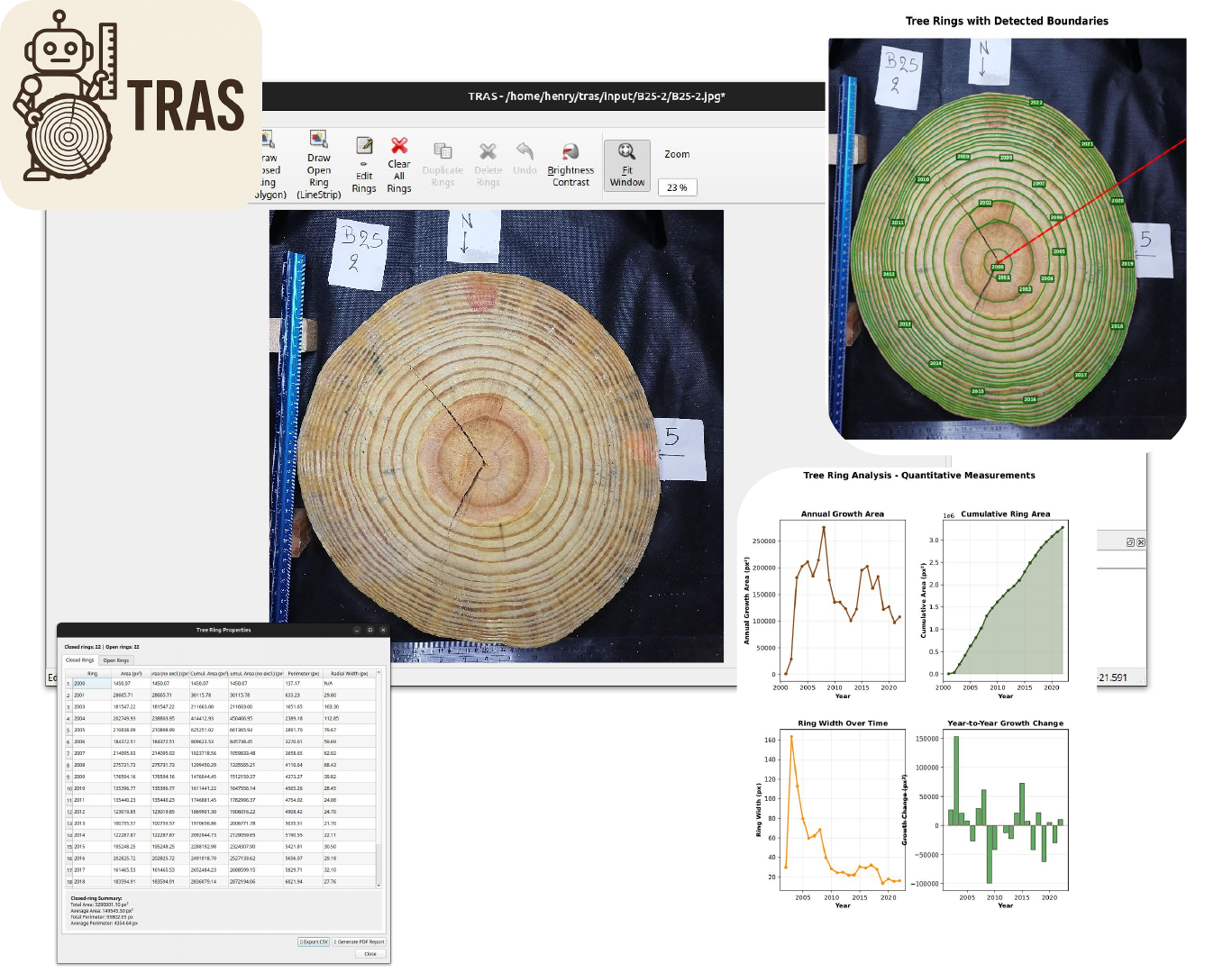}
    \caption{Overview of TRAS (Tree Ring Analyzer Suite). The figure illustrates the main components of the tool: automatic tree ring detection using CS-TRD, DeepCS-TRD, or INBD; interactive ring editing and correction via the PyQt5-based annotation canvas; and computation and export of dendrochronological metrics (area, perimeter, equivalent ring width). TRAS is available on Windows, macOS, and Linux.}
    \label{fig:tras_overview}
\end{figure}

In this work, we introduce the Tree Ring Analyzer Suite (TRAS), a tool with a graphical interface that is relatively easy to install, designed for the semi-automatic detection of growth rings in cross-sectional images with an option for manual correction. It incorporates the INBD \citep{inbd}, CS-TRD \citep{CSTRD}, and  DeepCS-TRD \citep{deepcstrd} automatic tree ring detection methods, which have been applied to various species, including the trees P. glauca, P. taeda, and G. triacanthos, as well as the shrubs \textit{Salix glauca L.}, \textit{D. octopetala}, \textit{E. hermaphroditum}, and \textit{V. myrtillus}.
The tool provides detailed two-dimensional measurements, facilitating an understanding of growth dynamics in tree species over time, and allows manual corrections of the automatically produced results. The semi-automatic character of TRAS significantly accelerates the measurement process of biological growth characteristics on a per-sample basis (see the results section). Moreover, ring measurements made by TRAS are  in this study compared with those of the CooRecorder tool \citep{Maxwell_2021}. An overview of the TRAS interface and its main components is shown in \Cref{fig:tras_overview}.

\section{Materials and Methods}

\subsection{Tree Ring Analyzer Suite}


The Tree Ring Analyzer Suite (TRAS) is an interactive software for tracing and editing tree rings in cross-sectional disc images. 
 
TRAS is implemented as a standalone desktop application built on top of a customized fork of \textit{LabelMe} \citep{wada2024labelme} using the PyQt5 framework. This design provides a robust, responsive, and flexible graphical environment that supports efficient visualization, editing, and correction of ring boundaries. The interface includes an enhanced annotation system, improved handling of high-resolution images, and streamlined tools for interactive editing, enabling a more efficient workflow for validating and adjusting automatic detections.

TRAS integrates state-of-the-art automatic ring detection algorithms, including CS-TRD, DeepCS-TRD, and INBD, which operate natively within the PyQt5 environment and interact seamlessly with the manual editing tools.

The tool is open source under the MIT license, and the code is available at \url{hmarichal93.github.io/tras}.

\textbf{Image menu} TRAS comprises four key steps. The first involves uploading the image, preprocessing it, and adding relevant metadata.  Once the acquired image is uploaded to the TRAS system, it undergoes several preprocessing steps to optimize the tree ring detection. Users can remove the background manually or automatically using a $U^{2}$Net network \citep{u2net}, which might improve the accuracy of the automated ring detection methods. 

\begin{figure}
    \centering
    \includegraphics[width=\linewidth]{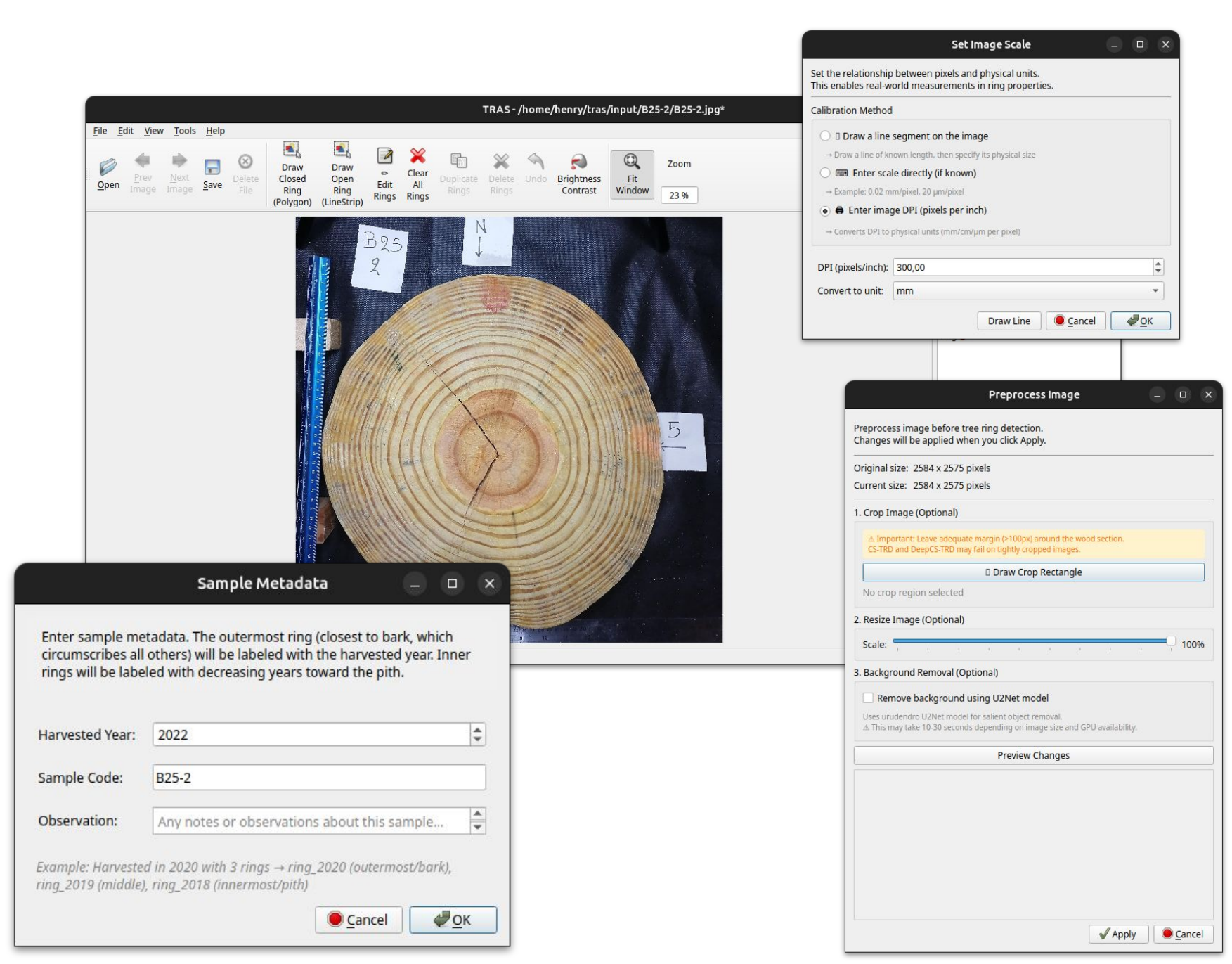}
    \caption{TRAS Image menu which allows preprocessing the image, adding relevant metadata, and setting the pixel-to-millimeter ratio}
    \label{fig:tras_image}
\end{figure}

Additionally, high-resolution images (greater than 10,000 pixels in width) can be resized to ensure smooth tool performance. To minimize interpolation artifacts during this process, Lanczos resampling is applied \citep{interplation}. A scale has to be set to establish the pixel-to-millimeter ratio required for accurate metric computations. \Cref{fig:tras_image} illustrates the menus for preprocessing the image, adding metadata, and setting the scale.

\begin{figure}
    \centering
    \includegraphics[width=0.7\linewidth]{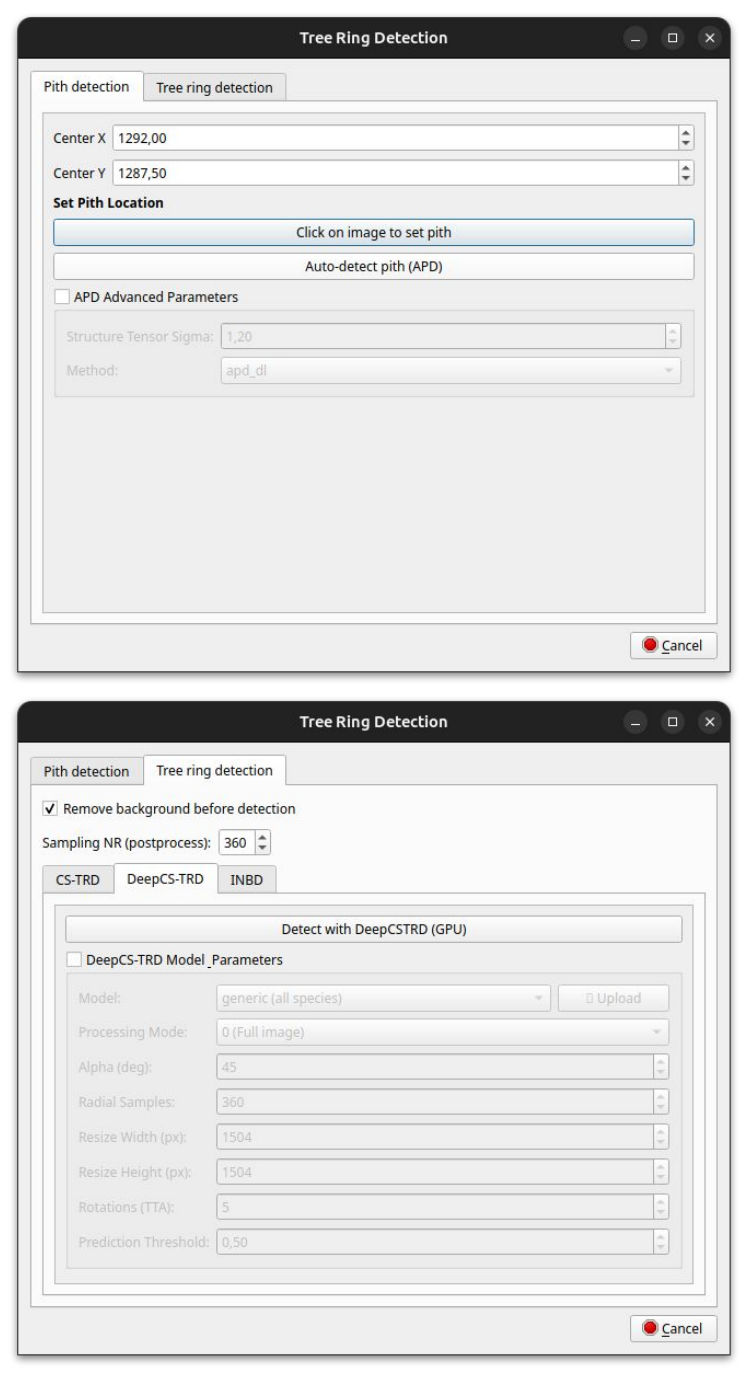}
    \caption{Menu for automatic tree ring detection (\textit{Ring Detection}).  The dialog is organized into two tabs. The \textit{Pith Detection} tab allows the user to define the pith location manually (by clicking on the image or entering coordinates) or automatically using one of three APD variants: APD, APD-PCL, or APD-DL.  The \textit{Tree Ring Detection} tab organizes CS-TRD, DeepCS-TRD, and INBD as separate method sub-tabs. Both deep learning-based methods allow users to upload a custom model or select from predefined models.  Shared options include background removal via U2Net and the number of radial sampling points used for post-processing ring resampling. Each method also exposes advanced parameters such as image resize dimensions; resulting ring boundaries are mapped back to the original image resolution.}
    \label{fig:tras_trd}
\end{figure}

\textbf{Ring Detection menu} The second step is the automatic detection of  tree rings (see \Cref{fig:tras_trd}). 
TRAS incorporates three algorithms for ring detection. . 
 
 The implemented methods are CS-TRD \citep{CSTRD}, a classical image processing approach based on edge detection and boundary recombination; DeepCS-TRD \citep{deepcstrd}, which replaces the edge detection stage of CS-TRD with a U-Net architecture; and INBD \citep{inbd}, a deep learning method that combines semantic segmentation with an iterative ring refinement stage.
A summary of the main characteristics of these algorithms, including architecture, training data, use of pretrained models, and reported performance, is provided in Table~\ref{tab:ring_detection_methods}. These automatic methods for ring detection need the location of the pith, and the first growth year. TRAS includes three automatic pith detector methods \citep{apd}: automatic wood pith detection (APD), PClines-based automatic wood pith detection (APD-PCL), and deep learning based automatic wood pith detection (APD-DL). Pith location can be marked manually if required as well. Additionally, the CS-TRD method can automatically detect the earlywood-latewood boundary if required.

\begin{sidewaystable}
\centering
\small
\caption{Summary of the ring detection algorithms implemented in TRAS. The reported performance values correspond to the UruDendro4 benchmark \citep{urudendro4}, which evaluates all methods on 102 RGB cross-sections of \textit{Pinus taeda}.}
\label{tab:ring_detection_methods}
\begin{tabular}{p{1.5cm}|p{1.8cm}|p{2.7cm}|p{3.2cm}|p{2.6cm}|p{1.4cm}|p{2.0cm}}
\hline
Algorithm & Type & Main idea / \newline architecture & Training / development \newline data & Benchmark \newline data & Pretraining & Reported \newline performance \\ 
\hline
CS-TRD & Classical image processing & Edge detection and boundary recombination & No training required; development based on 64 RGB cross-sections of \textit{P. taeda} & UruDendro4: 102 RGB cross-sections of \textit{P. taeda} & None & mAR = .568, \newline ARAND = .174 \\ 
\hline
INBD & Deep learning + postprocessing & U-Net for segmentation + \newline iterative ring refinement & 213 microscopy images from \textit{D. octopetala}, \textit{E. hermaphroditum}, and \textit{V. myrtillus} & UruDendro4: 102 RGB cross-sections of \textit{P. taeda} & ImageNet-1K & mAR = .712, \newline ARAND = .099 \\ 
\hline
DeepCS-TRD & Deep learning + postprocessing & U-Net for boundary detection + \newline CS-TRD postprocessing & 117 RGB cross-sections of \textit{P. taeda} and 9 of \textit{G. triacanthos} & UruDendro4: 102 RGB cross-sections of \textit{P. taeda} & ImageNet-1K & mAR = .775, \newline ARAND = .087 \\
\hline
\end{tabular}
\end{sidewaystable}

\begin{figure}
    \centering
    \includegraphics[width=\linewidth]{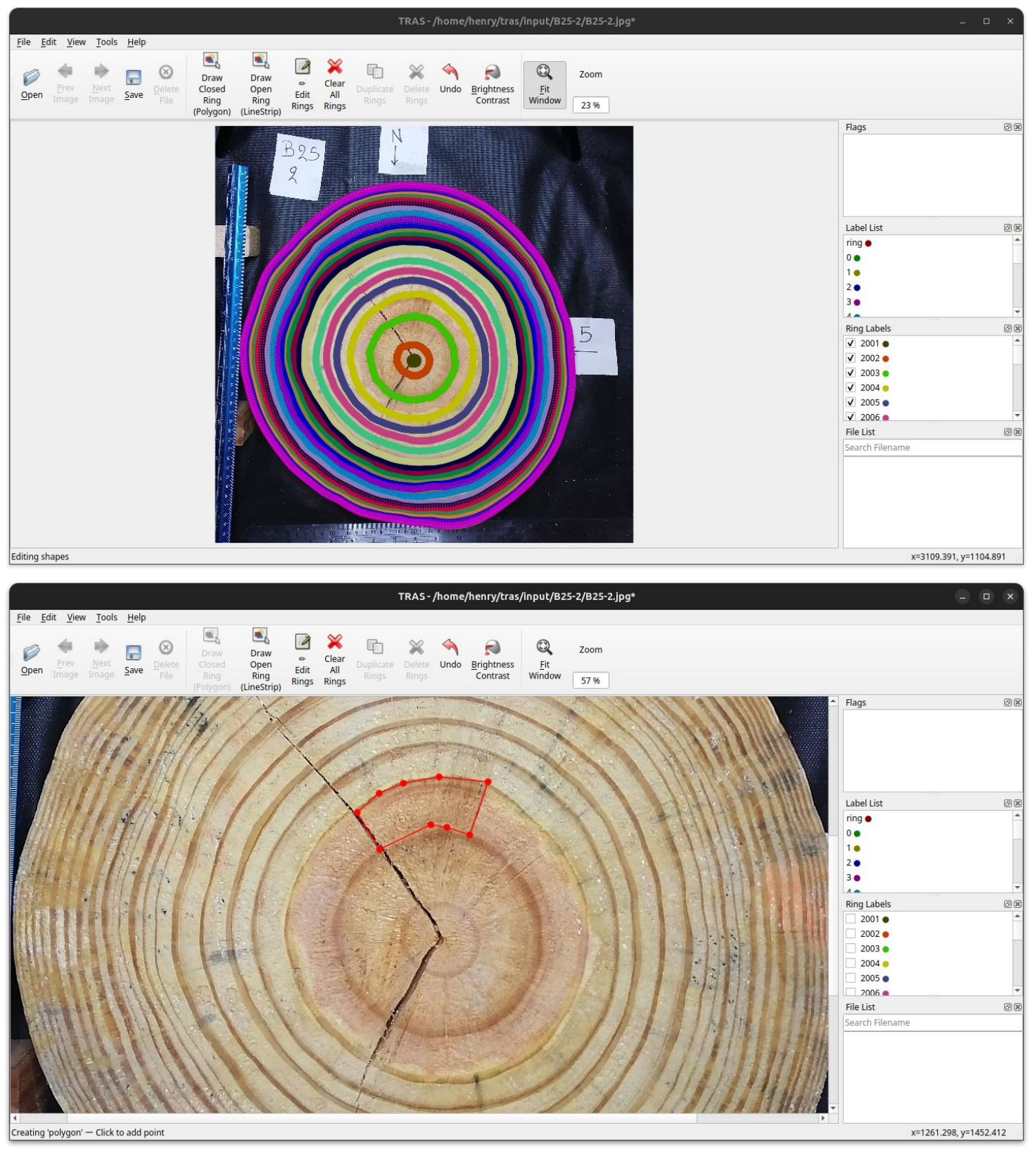}
    \caption{Menu for automatic tree ring postprocessing (\textit{Ring Editing}). Rings are represented as polylines with a fixed number of points, which can be moved, deleted, or added as needed. Each polyline is automatically labeled with the corresponding growth year, based on metadata provided (e.g., harvest year).  Additionally, other shapes such as cracks, knots, or regions of interest can be annotated as separate polygons; their area can then be excluded from ring computations. In the figure, a region of interest is highlighted with a red polyline.}
    \label{fig:tras_postprocessing}
\end{figure}

\textbf{Ring Editing menu} The third step is the interactive tree ring adjustment. A Python-based graphical user interface  built with PyQt5 allows user interaction. Predictions generated by the automated method can be refined using  a LabelMe-based \citep{wada2024labelme}  annotation canvas integrated directly into the application. This setup allows users to remove false ring detections, correct existing ones, and add any rings not initially detected. The same interface will enable users to manually delineate the tree rings without using automatic tools or delineate the \textit{earlywood-latewood} boundaries. Other shapes, such as cracks or knots, can be manually delineated. \Cref{fig:tras_postprocessing} illustrates the editing panel.

The correction workflow involves three main operations. To \textit{delete} a false detection, the user clicks the ring boundary to select it and presses \texttt{Delete}. To \textit{edit} an existing ring, the user enters Edit mode (\texttt{Ctrl+J}), then drags individual polyline points to the correct position; additional points can be inserted by \texttt{Alt}-clicking on any edge segment, and unwanted points removed by selecting them and pressing \texttt{Backspace}. To \textit{add} a missing ring, the user draws a new polyline by pressing \texttt{Ctrl+N} (closed polygon) or \texttt{Ctrl+Shift+N} (open linestrip), clicking to define each boundary point, and pressing \texttt{Enter} to finish. All operations support undo (\texttt{Ctrl+Z}). A complete list of keyboard shortcuts is provided in Appendix \ref{sec:shortcuts}.

\begin{figure}
    \centering
    \includegraphics[width=\linewidth]{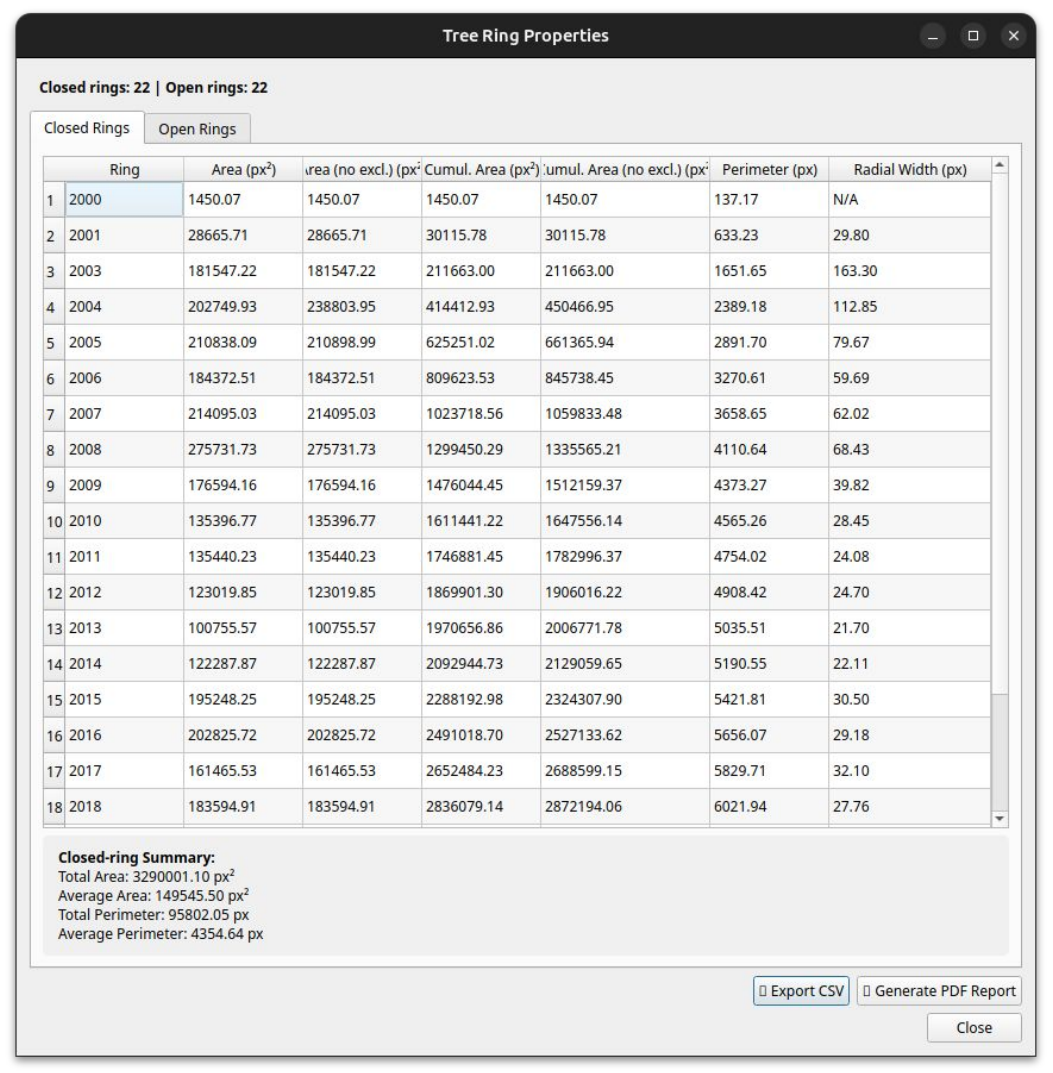}
    \caption{ Ring Properties dialog displaying computed metrics in tabular format. For each annotated ring, the table reports area, perimeter, equivalent ring width, eccentricity, and similarity factor.}
    \label{fig:tras_visualization}
\end{figure}

\textbf{Metrics menu} Finally, the user can compute various metrics (such as area, perimeter, etc.) related to the annual rings or to the \textit{earlywood–latewood} boundaries. The option to delineate the \textit{earlywood–latewood} boundary is included because several studies have shown that latewood ring width is a better proxy for climate variability than total ring width \citep{latewood_1,latewood_2}. To compute the ring properties, such as area or perimeter, from the ring boundary points, the Shapely library \citep{shapely} is used.
\Cref{fig:tras_visualization} displays the Metrics menu, where the data can be presented in tabular or graphical format (export pdf option) for a more straightforward analysis.

\begin{figure}
\centering
   \includegraphics[width=\linewidth]{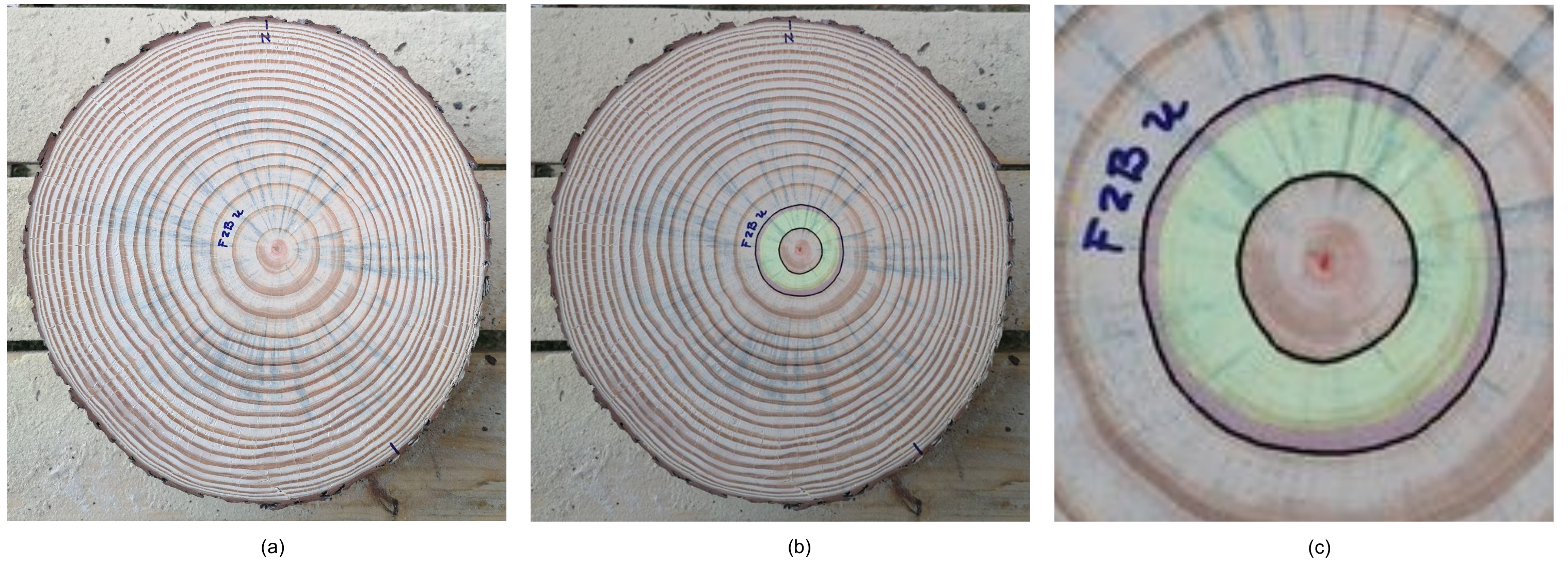}
   \caption{Ring structures. The cumulative EarlyWood (EW) area is equal to  Area(Ring) + Area(EW). The cumulative ring area is equal to Area(Ring) + Area(EW) + Area(LW). (a) Cross-section image (b) Cross-section image where Area(EW) is highlighted in green while Area(LW) is highlighted in light blue. Ring boundaries are highlighted in black. (c) Zoomed-in image.}
   \label{fig:cumulative}
\end{figure}

Several indicators can be computed, which include area-based metrics such as \textit{earlywood (EW) area},\textit{ latewood (LW) area},\textit{ cumulative EW area}, and \textit{cumulative ring area}.~\Cref{fig:cumulative} illustrates the concept of cumulative areas. If additional shapes such as cracks, knots, or fungus are delineated, the tool provides an option to compute the ring area while excluding the area occupied by these shapes. For instance,  the area of the region of interest highlighted in \Cref{fig:tras_postprocessing}  (red polyline) can be subtracted from the ring-growth area as needed.

In addition, the perimeters of EW and LW (annual ring) can be computed. The \textit{equivalent ring width} is defined as 
$$\Delta r^{eq}_i = \sqrt{Area_{i}/\pi} - \sqrt{Area_{i-1}/\pi}$$
where $Area_{i}$ is the area of \textit{ring i} and $\Delta r^{eq}_i$ is the \textit{equivalent ring width} of ring $i$. We use a hypothetical circle with the same area as the ring $i$.

A \textit{Ring's Circle Similarity Factor} (SF) is also added. This indicator measures how far the ring is from a perfect circle (takes values between 0 and 1). It is computed as $$SF=1 - \frac{Perimeter_i - 2\sqrt{\pi Area_i}}{Perimeter_i}$$ where $Perimeter_i$ and $Area_i$ are the perimeter and area of \textit{ring $i$}. 

The \textit{Ring Eccentricity} is measured by the difference between the centroid of the cumulative area of the annual ring and the pith. It is a vector; therefore, the module and phase of the eccentricity are reported.

\begin{figure}
    \centering
    \includegraphics[width=\linewidth]{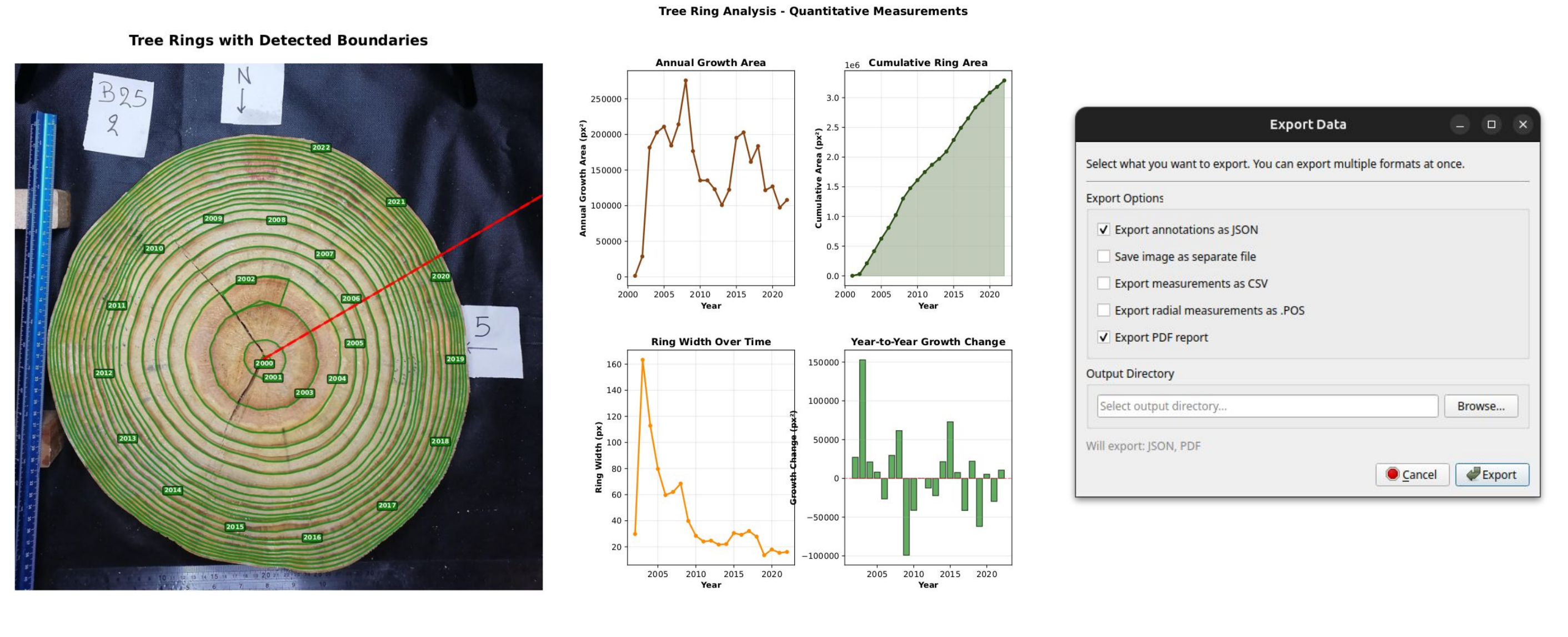}
    \caption{Export outputs of TRAS. (a) PDF report page showing ring boundaries overlaid on the cross-section image with the radial measurement ray. (b) PDF report page with analysis plots of the computed ring metrics. (c) Export dialog, where the user selects the desired output formats (PDF, CSV, \textit{pos}) and configures export options.}
    \label{fig:tras_export}
\end{figure}

 TRAS also provides a dedicated \textit{Measure Ring Width} dialog that allows the user to draw a radial ray from the pith; ring widths are then computed as the distances between consecutive intersections of that ray with each ring boundary, following a similar approach to \citep{Maxwell_2021}.  Results can be exported in \textit{pos}, CSV, and PDF formats. The PDF report includes ring boundary overlays with the measurement ray as well as analysis plots of the computed metrics, as illustrated in \Cref{fig:tras_export}.

\subsection{Installation and usage}

TRAS is distributed as an open-source Python package hosted at \url{https://github.com/hmarichal93/tras}. Installation requires Anaconda and proceeds in four steps: (1) download the source archive from the repository releases page and extract it; (2) create and activate the conda environment (\texttt{conda env create -f environment.yml \&\& conda activate tras}); (3) install the package (\texttt{pip install -e .}); and (4) download the required model assets \\(\texttt{python tools/download\_release\_assets.py}). Optionally, CS-TRD requires compiling the Devernay edge detector (Linux/macOS only), and INBD requires cloning a separate repository; both steps are detailed in the project documentation.

The graphical interface is launched from the terminal with \texttt{tras}, optionally followed by an image path. The CLI tool \texttt{tras\_detect} supports single-image mode (\texttt{tras\_detect image.jpg -o output.json}) and batch-folder mode, the latter accepting a YAML configuration file for scale, preprocessing, and detection parameters. Batch outputs include a JSON annotation file, a CSV metrics table, and a PDF report per image.

\subsection{Datasets}

\subsubsection*{\textit{P. taeda} annotated cross sections}
\label{sec:dataset}

\begin{figure}
    \centering
    \includegraphics[width=\linewidth]{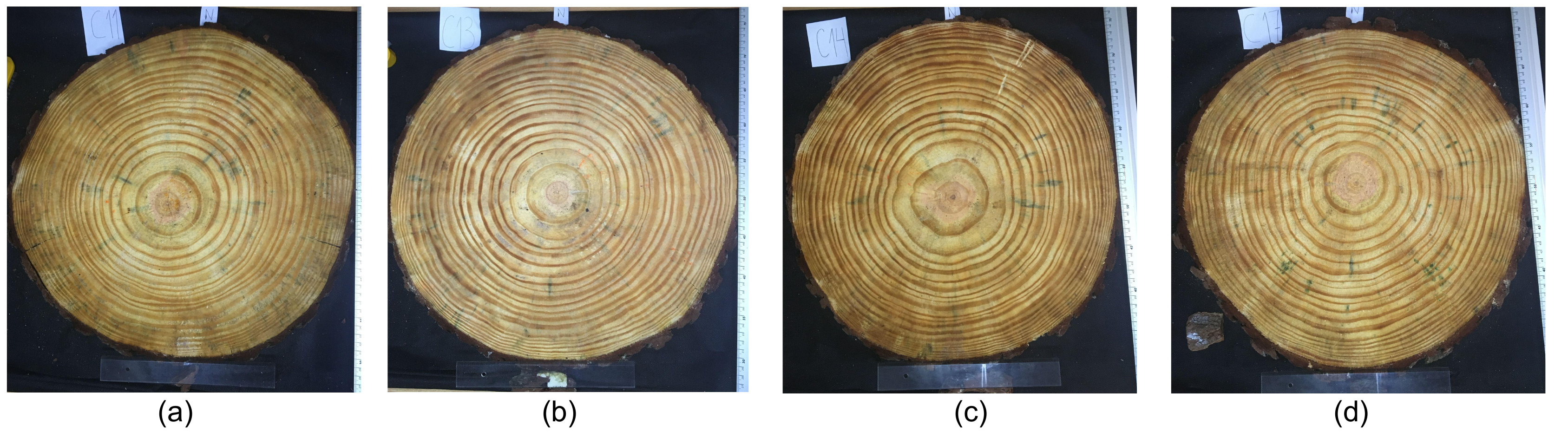}
    \caption{\textit{Pinus taeda} discs images. (a) C11C. (b) C13C. (c) C14C. (d) C17C.}
    \label{fig:four_images}
\end{figure}

The dataset used to assess the automatic tree-ring detection accuracy and measurements consists of 18 RGB images of \textit{P. taeda} cross-sections (see \Cref{fig:four_images} for examples). The images were taken in a controlled laboratory setting with an iPhone 6S (12 MP camera) positioned 43 to 51 cm from the sample and illuminated by a 35 W LED ring. Image dimensions range from 1000 to 3000 pixels in width. The cross-sections exhibit different surface finishing: some were cut with a chainsaw, others were smoothed with a handheld planer, and some were polished with a rotary sander. The cross-sections were extracted from 23-year-old trees.

Several experts manually traced annual tree-ring curves on each image sample. One expert annotated each image, a second reviewed and adjusted the annotations as necessary, and a third conducted a final review. This iterative approach ensured high accuracy, making these annotations reliable ground truths for evaluating machine learning models. 

Manual measurements of annual ring widths were taken along the north, south, east, and west directions for the disc samples in~\Cref{fig:four_images}. These measurements are a reference for validating our tool's performance and the outputs from CooRecorder \citep{Maxwell_2021}.

\subsubsection*{External datasets}

Additionally, we applied TRAS to images from external datasets that presented additional challenges, such as cracks, fungal damage, or atypical ring patterns, to validate the tool further. In all cases, ring boundary curves were generated using TRAS:

\begin{itemize}
    \item \cite{marichal2023urudendro}: Samples F03d and L02c from the species P. taeda (see \Cref{fig:f03d,fig:l02b}).
    \item \cite{deepcstrd}: Sample G6 from the species G. triacanthos (see \Cref{fig:g6}).
    \item \cite{salix}: Sample 1W23S20 from the species S. glauca (see \Cref{fig:salix}).
    \item \cite{DouglasFir}: Sample C07c from the species Douglas fir (see \Cref{fig:douglas}).
\end{itemize}

\subsection{Validation and accuracy assessment}
\label{sec:validation_and_acc}
\subsubsection*{Automatic tree ring detection}

To quantify the manual effort required from the user to obtain complete and accurate ring delineations, -i.e., adding, editing, or deleting rings- we compared the detections produced by the methods integrated in TRAS against expert annotations, as described in the dataset section. For the neural network–based methods INBD and DeepCS-TRD, we used the models presented in \cite{deepcstrd}, which were trained on the P. taeda image dataset introduced in \cite{marichal2023urudendro}. The CS-TRD method is used with the default parameters.

We assessed the performance of the automatic ring detection methods by evaluating the number of annual rings correctly identified in the P. taeda dataset. All detections were performed through the TRAS web interface. A resize factor of 2 was applied to each image to enhance performance, and the background was automatically removed. The pith location was provided by the APD automatic detection method \citep{apd}.

As evaluation metrics for the automatic detections, we use \textit{Precision}, \textit{Recall}, and the \textit{F-Score}, as defined in \cite{marichal2023urudendro}. Precision is calculated as $P=\frac{TP}{TP+FP}$, Recall as $R=\frac{TP}{TP+FN}$, and the F-Score as $F=\frac{2P \times R}{P+R}$, where $TP$ denotes true positives, rings correctly detected; $FP$ denotes false positives, detections that do not correspond to any ring; and $FN$ denotes false negatives, rings that were not detected.

A distance-based matching determines whether each automatically generated ring ($d_i$) corresponds to a ground truth ring ($g_i$). For each ground truth ring, the closest detection (\textit{dt}) is assigned if more than 90\% (fixed value) of its nodes lie within the area of influence of the \textit{gt} ring. This area of influence is defined as the band between half the distance to each neighboring ring, effectively encompassing the set of pixels closest to \textit{gt}. All unassigned automatic detections are counted as false positives ($FP$).

\subsubsection*{Measurements}

One of the main features of TRAS is the computation of two-dimensional ring measurements, such as ring area. However, to the best of our knowledge, there is currently no widely used software that provides directly comparable area measurements from cross-sectional images, so this output could not be externally validated.

Therefore, the comparison with CooRecorder \citep{Maxwell_2021} focused on ring width measurements only. CooRecorder is a well-established tool for measuring tree-ring widths along user-defined segments, and TRAS implements a comparable procedure. In both cases, measurements were derived from the same post-processed ring delineations. Agreement between the two tools was assessed by comparing the resulting width measurements using the Pearson correlation coefficient.

This analysis was performed on four disc samples (C17C, C13C, C11C, and C14C), considering four directional paths per sample. Along each path, 21 ring widths were measured with both tools, resulting in 336 paired observations.

\section{Results}
\label{sec:results}

\subsection{Automatic tree ring detection}

\begin{table*}
\centering
\caption{Automatic tree ring detection in the \textit{P. taeda} dataset with the INBD, CS-TRD, and DeepCS-TRD methods. Each row presents the fully automatic ring prediction for each sample. As columns, the metrics Precision (\textbf{P}), Recall (\textbf{R}), and F-Score (\textbf{F}) are presented. The last row represents the average per column.}
\label{tab:results_all_dataset}
\begin{tabular}{l|ccc|ccc|ccc}
        & \multicolumn{3}{c|}{INBD} & \multicolumn{3}{c|}{CS-TRD} & \multicolumn{3}{c}{DeepCS-TRD} \\ \hline
Sample  & P      & R      & F      & P       & R       & F      & P        & R        & F        \\ \hline 
B3C     & 76.2   &   69.6     & 72.7   & 76.2    & 69.6    & 72.7   & 90.0     & 78.3     & 83.7     \\
B19C    & 81.8   & 78.3   & 80.0   & 94.7    & 78.3    & 85.7   & 90.0     & 78.3     & 83.7     \\
B2C     & 66.7   & 60.9   & 63.6   & 72.2    & 56.5    & 63.4   & 73.7     & 60.9     & 66.7     \\
A29C    & 13.0   & 13.0   & 13.0   & 90.0    & 78.3    & 83.7   & 90.0     & 78.3     & 83.7     \\
C10C    & 72.7   & 69.6   & 71.1   & 66.7    & 52.2    & 58.5   & 84.2     & 69.6     & 76.2     \\
A10C    & 95.5   & 91.3   & 93.3   & 80.0    & 69.6    & 74.4   & 95.0     & 82.6     & 88.4     \\
C11C    & 85.7   & 78.3   & 81.8   & 84.2    & 69.6    & 76.2   & 70.0     & 60.9     & 65.1     \\
C13C    & 95.2   & 87.0   & 90.9   & 86.4    & 82.6    & 84.4   & 95.0     & 82.6     & 88.4     \\
B24C    & 90.9   & 87.0   & 88.9   & 94.4    & 73.9    & 82.9   & 90.0     & 78.3     & 83.7     \\
A4C     & 77.3   & 94.4   & 85.0   & 81.8    & 100.0   & 90.0   & 81.0     & 94.4     & 87.2     \\
A30C    & 95.5   & 91.3   & 93.3   & 90.9    & 87.0    & 88.9   & 90.5     & 82.6     & 86.4     \\
C14C    & 77.3   & 73.9   & 75.6   & 85.0    & 73.9    & 79.1   & 81.0     & 73.9     & 77.3     \\
A7C     & 90.5   & 82.6   & 86.4   & 94.4    & 73.9    & 82.9   & 75.0     & 65.2     & 69.8     \\
B25C    & 91.3   & 91.3   & 91.3   & 86.4    & 82.6    & 84.4   & 90.5     & 82.6     & 86.4     \\
B12C    & 90.5   & 82.6   & 86.4   & 100.0   & 78.3    & 87.8   & 89.5     & 73.9     & 81.0     \\
A9C     & 76.2   & 69.6   & 72.7   & 75.0    & 78.3    & 76.6   & 85.0     & 73.9     & 79.1     \\
C16C    & 95.5   & 91.3   & 93.3   & 85.0    & 73.9    & 79.1   & 90.5     & 82.6     & 86.4     \\
C17C    & 90.5   & 82.6   & 86.4   & 90.0    & 78.3    & 83.7   & 94.7     & 78.3     & 85.7     \\ \hline
Average & 81.2   & 77.5   & 79.2   & 85.2    & 75.4    & 79.7   & 86.4     & 76.5     & 81.0    
\end{tabular}
\end{table*}

\Cref{tab:results_all_dataset} presents the average performance of automatic ring detection across 18 \textit{P. taeda} samples (comprising 414 tree rings), using the TRAS web interface. All detections were performed via the \textit{Ring Detection} menu (see \Cref{fig:tras_trd}), without any user postprocessing. On average, the three methods showed comparable performance. DeepCS-TRD achieved the highest precision at 86.4\%, while INBD recorded the lowest at 81.2\%. Regarding recall, INBD outperformed the others with 77.5\%, whereas CS-TRD obtained the lowest value at 75.4\%. Regarding the F-Score, DeepCS-TRD again achieved the best result with 81.0\%, while INBD reached 79.2\%.

Execution times for automatic detection varied across methods. CS-TRD completed processing in an average of 21.0 seconds per sample, INBD required 45.7 seconds, and DeepCS-TRD 66.6 seconds. These measurements were obtained on a workstation with an Intel Core i5-10300H processor and 16 GB of RAM (CPU only). When using a GPU (NVIDIA GTX 1650), INBD achieved a significantly faster runtime of 6.7 seconds per image, while DeepCS-TRD required 29.5 seconds.



\subsection{Postprocessing Examples}

\begin{figure}
    \centering
    \includegraphics[width=\linewidth]{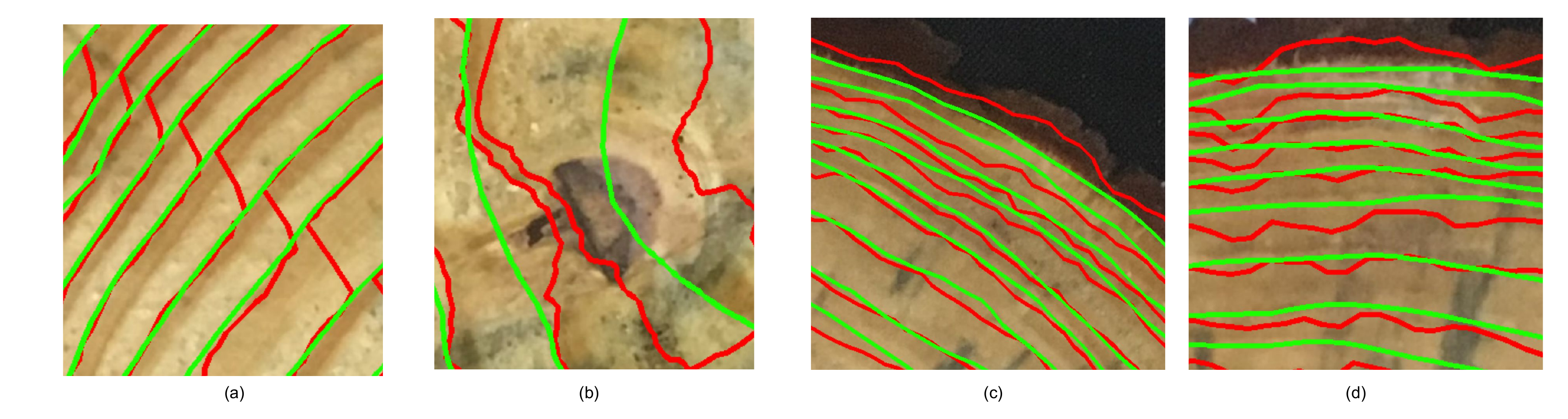}
    \caption{Sample B3C. The automatic ring detection is drawn in red, and the results are in green after being corrected by an expert. Some typical errors: a) Detection ring jump propagation. b) Knot ring-induced errors. c) and d) lack of precision in the detection }
    \label{fig:postprocessing}
\end{figure}

\Cref{fig:postprocessing} illustrates some cases that required manual postprocessing. The initial automatic ring detection is shown in red, while the corrected detection by an expert is shown in green. A typical error in the INBD method is the propagation of detection errors. Since this method detects rings sequentially from the pith to the bark, a detection error can propagate iteratively to outer rings, as shown in \Cref{fig:postprocessing}a. Knots may sometimes interfere with the automatic method, leading to false ring detections (\Cref{fig:postprocessing}b). In some cases, the automatic detections lack precision, as seen in~\Cref{fig:postprocessing}c and~\Cref {fig:postprocessing}d.

\subsection{Measurements}

To evaluate the measurement capabilities of TRAS, we compared its ring-width measurements with those obtained using CooRecorder \citep{Maxwell_2021}, a widely used tool for one-dimensional tree-ring analysis. This comparison was performed on the same post-processed ring delineations, that is, after manual user corrections had been applied to the automatically detected ring boundaries in TRAS. Therefore, this experiment assesses the agreement between the measurement procedures of both tools rather than the performance of the automatic ring-detection algorithms.

The analysis included four disc samples (C17C, C13C, C11C, and C14C). For each sample, ring widths were measured along four directional paths (north, south, east, and west), and 21 consecutive rings were measured per path with both tools, yielding a total of 336 paired observations.

\begin{figure}
    \centering
    \includegraphics[width=\linewidth]{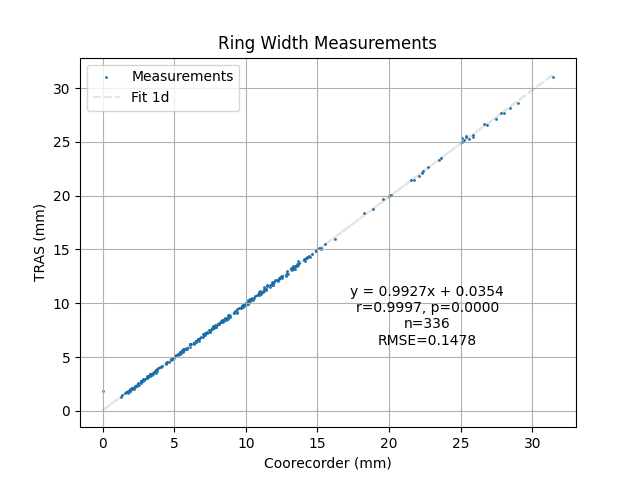}
    \caption{Comparison between tree ring width measurements with CooRecorder and TRAS. A linear correlation can be distinguished}
    \label{fig:coorecorder}
\end{figure}

\Cref{fig:coorecorder} shows the relationship between the ring-width values produced by TRAS and CooRecorder. The agreement between both tools was very high (Pearson's $r > 0.99$, $p < 0.001$, $n = 336$). The fitted linear regression had a slope of 0.9927 and an intercept of 0.0354, indicating an almost one-to-one correspondence between the two measurement procedures. The root mean square error between paired measurements was 0.1478\,mm.

Beyond conventional path-based ring-width measurements, TRAS also computes \\two-dimensional descriptors from the corrected ring polygons, including  annual ring area, cumulative ring area, annual-ring perimeter, earlywood area, latewood area, cumulative earlywood area, equivalent ring width, ring eccentricity, and circle similarity factor (\Cref{fig:tras_visualization,fig:cumulative}). These metrics constitute one of the main contributions of the software, as they enable the quantitative characterization of asymmetric radial growth in cross-sectional images.  An example of the tabular export generated by TRAS is shown in \Cref{tab:tras_area_export}. For each ring, the software reports the individual ring area, cumulative ring area, and ring perimeter in physical units and in pixels. In addition, TRAS automatically generates a PDF report that summarizes the analyzed sample and includes graphical representations of annual growth area, cumulative ring area, area distribution, and year-to-year growth change. These outputs provide complementary information to conventional radial ring-width series and facilitate the inspection of growth patterns directly from cross-sectional images. Since no established software currently provides directly comparable area-based measurements for this type of data, these outputs could not be externally validated in the same way as ring width. Nevertheless, they represent a central functionality of TRAS and extend tree-ring analysis beyond one-dimensional radial measurements.

\begin{table}[t]
\centering
\caption{Example of area-based measurements exported by TRAS for consecutive rings of one sample.}
\label{tab:tras_area_export}
\resizebox{\columnwidth}{!}{%
\begin{tabular}{cccc}
\hline
Ring & Area (mm$^2$) & Cumul.\ area (mm$^2$) & Perimeter (mm) \\
\hline
0  & 43.07   & 43.07    & 23.64  \\
1  & 851.53  & 894.60   & 109.14 \\
2  & 5392.93 & 6287.53  & 284.67 \\
3  & 7093.76 & 13381.28 & 411.78 \\
4  & 6264.83 & 19646.11 & 498.39 \\
5  & 5476.85 & 25122.97 & 563.70 \\
\hline
\end{tabular}%
}
\end{table}

\Cref{fig:coorecorder} 

\section{Discussion}

\subsection{Performance of Automatic Detection Methods}

The results presented in \Cref{tab:results_all_dataset} demonstrate that all three integrated detection algorithms (INBD, CS-TRD, and DeepCS-TRD) achieve satisfactory performance in automatic tree ring detection for \textit{P.\ taeda} samples. DeepCS-TRD obtained the highest F-score (81.0\%) and precision (86.4\%), indicating it produces fewer false positives compared to the other methods. This superior precision is particularly valuable in practical applications, as it minimizes the manual effort required to remove incorrect detections. The higher recall achieved by INBD (77.5\%) suggests this method is more effective at detecting faint or ambiguous ring boundaries, though at the cost of introducing more false positives. The false positives are not really a problem, as they can be easily removed by the user.

The comparable performance across methods indicates that TRAS provides users with multiple algorithmic options to address different image characteristics and quality variations. For instance, CS-TRD, despite being a classical image processing approach, achieved competitive results while requiring the shortest processing time (21.0 seconds on CPU), making it suitable for rapid analysis when computational resources are limited. The deep learning methods, particularly when GPU acceleration is available, offer improved accuracy with acceptable processing times (6.7 seconds for INBD and 29.5 seconds for DeepCS-TRD with GPU).

It is important to note that the relatively modest recall values (75.4\% to 77.5\%) across all methods reflect the inherent challenges in automatic tree ring detection, including variations in wood surface quality, presence of knots and cracks, and subtle ring boundaries in certain growth periods. However, these results align with the semi-automatic nature of TRAS, where the graphical interface enables manual correction of missed detections.

\subsection{Common Detection Errors and Manual Correction}

The analysis of detection errors illustrated in \Cref{fig:postprocessing} reveals several systematic challenges that affect automatic ring detection. The propagation of detection errors observed with the INBD method (\Cref{fig:postprocessing}a) is an inherent limitation of sequential detection approaches. Since INBD processes rings iteratively from pith to bark, an error in detecting an inner ring can cascade to outer rings, requiring users to correct multiple boundaries. This behavior contrasts with CS-TRD and DeepCS-TRD, which detect all rings simultaneously, thereby avoiding error propagation.

Knots represent a significant source of false positives across all methods (\Cref{fig:postprocessing}b). The anatomical structure of knots creates density variations and edge patterns that mimic ring boundaries, particularly in images with lower surface quality. The ability to manually delineate and exclude knot regions through TRAS's interface (as demonstrated in \Cref{fig:tras_postprocessing}) provides a practical solution to this challenge, allowing users to compute ring areas while explicitly accounting for these anomalies.

The precision issues shown in \Cref{fig:postprocessing}c and d, where automatic detections deviate from true ring boundaries, are common in regions with compressed or diffuse latewood-earlywood transitions. The postprocessing interface in TRAS addresses these limitations by allowing users to adjust individual points along ring boundaries, ensuring accurate measurements without requiring complete re-annotation.

\subsection{Validation of TRAS Measurements}

The strong correlation between TRAS and CooRecorder measurements (Pearson's $r > 0.99$, \Cref{fig:coorecorder}) validates the accuracy of TRAS's ring width computation along user-defined paths. The near-unity slope (0.9927) and near-zero intercept (0.0354) of the linear regression indicate that both tools produce equivalent measurements when applied to the same corrected ring delineations. The Root Mean Square Error of 0.1478\,mm falls within acceptable limits for dendrochronological studies, particularly considering that measurement variability can arise from differences in path positioning or interpolation methods between tools.

This validation is crucial because it demonstrates that TRAS can serve as a reliable alternative to established tools for traditional ring width measurements while simultaneously offering advanced capabilities for area-based metrics. The ability to compute equivalent ring width, cumulative areas, and shape factors (as illustrated in \Cref{fig:tras_visualization} and \Cref{fig:cumulative}) positions TRAS as a comprehensive solution for diverse dendrochronological applications.

\subsection{Generalization Across Species and Image Conditions}

\begin{figure}
    \centering
    \includegraphics[width=\linewidth]{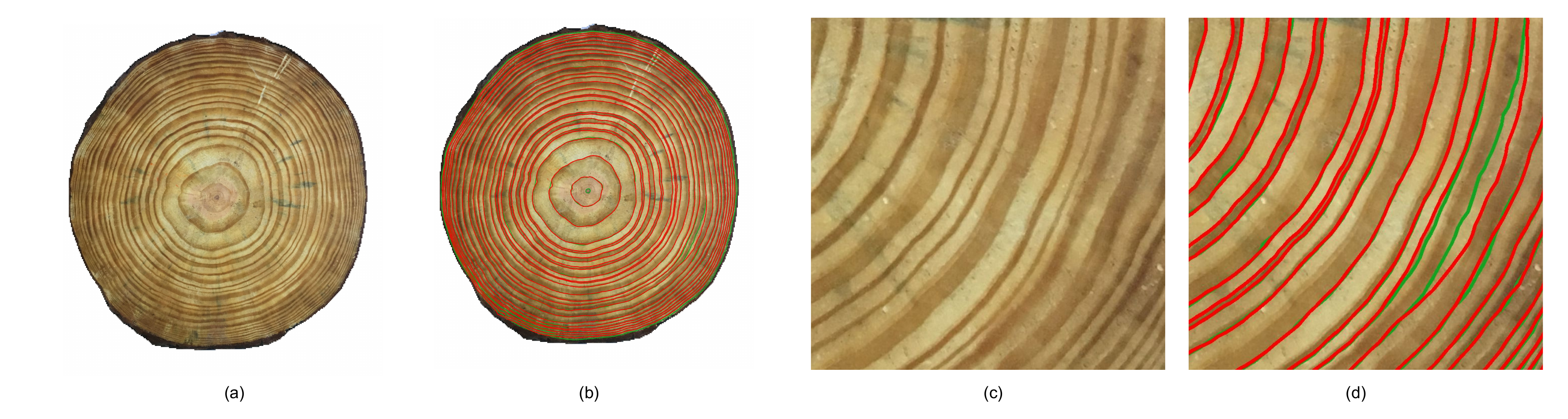}
    \caption{P. taeda sample C14C. (a) Sample after background removal. (b) Expert-delineated rings are shown in green, and automatically detected rings are shown in red. Correct detections overlap with the expert annotations. (c, d) Zoomed-in views highlighting a disk's region.}
    \label{fig:c14}
\end{figure}

\begin{figure}
    \centering
    \includegraphics[width=\linewidth]{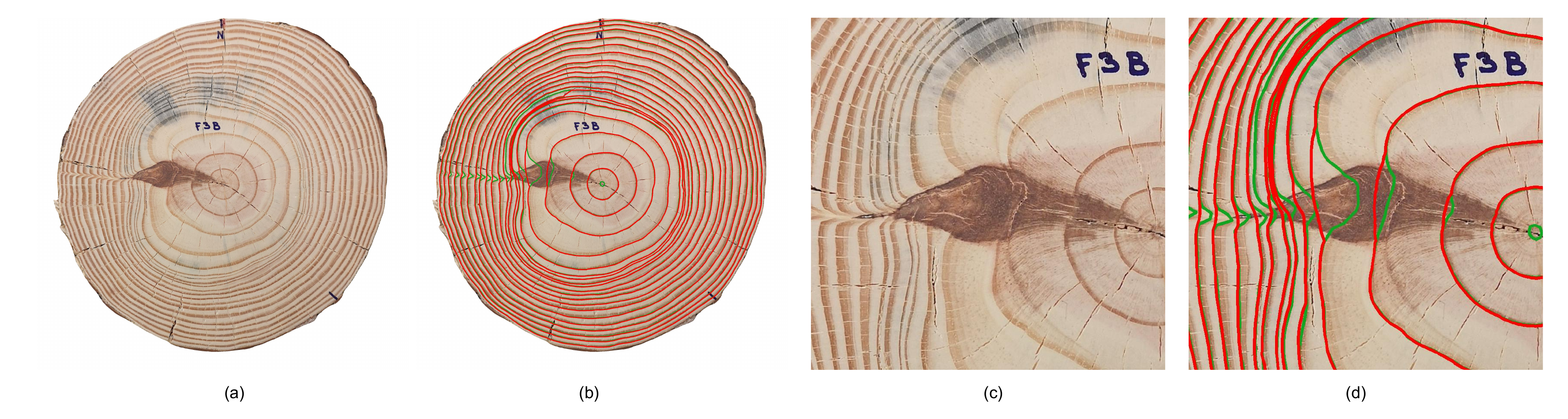}
    \caption{P. taeda image sample F03d from \cite{marichal2023urudendro}. (a) Sample after background removal. (b) Expert-delineated rings are shown in green, and automatically detected rings are shown in red. Correct detections overlap with the expert annotations. (c, d) Zoomed-in views highlighting a disk's region.}
    \label{fig:f03d}
\end{figure}

\begin{figure}
    \centering
    \includegraphics[width=\linewidth]{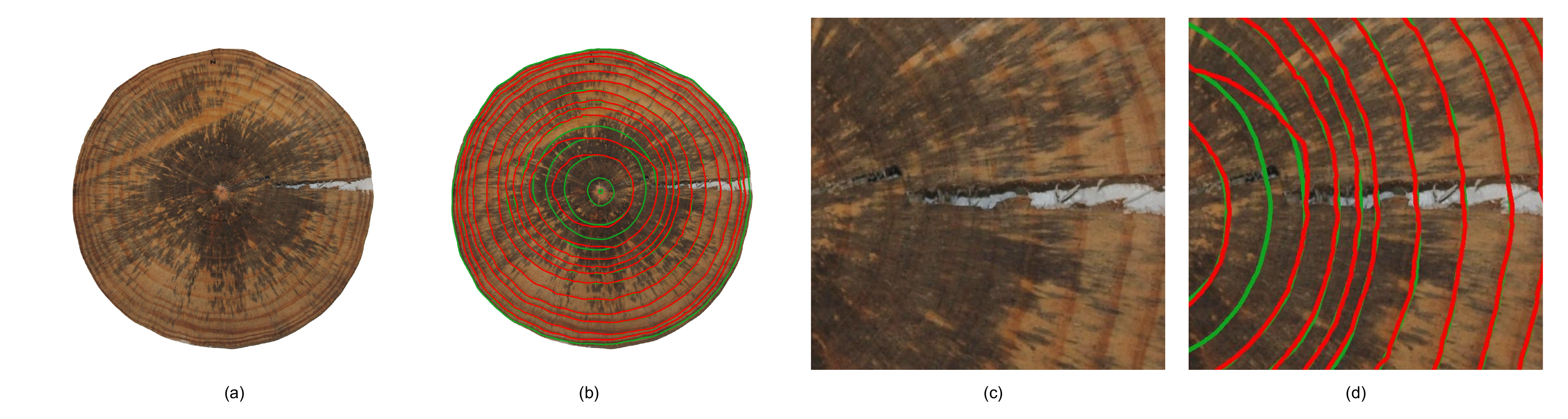}
    \caption{P. taeda image sample L02b from \cite{marichal2023urudendro}. (a) Sample after background removal. (b) Expert-delineated rings are shown in green, and automatically detected rings are shown in red. Correct detections overlap with the expert annotations. (c, d) Zoomed-in views highlighting a disk's region. }
    \label{fig:l02b}
\end{figure}

\begin{figure}
    \centering
    \includegraphics[width=\linewidth]{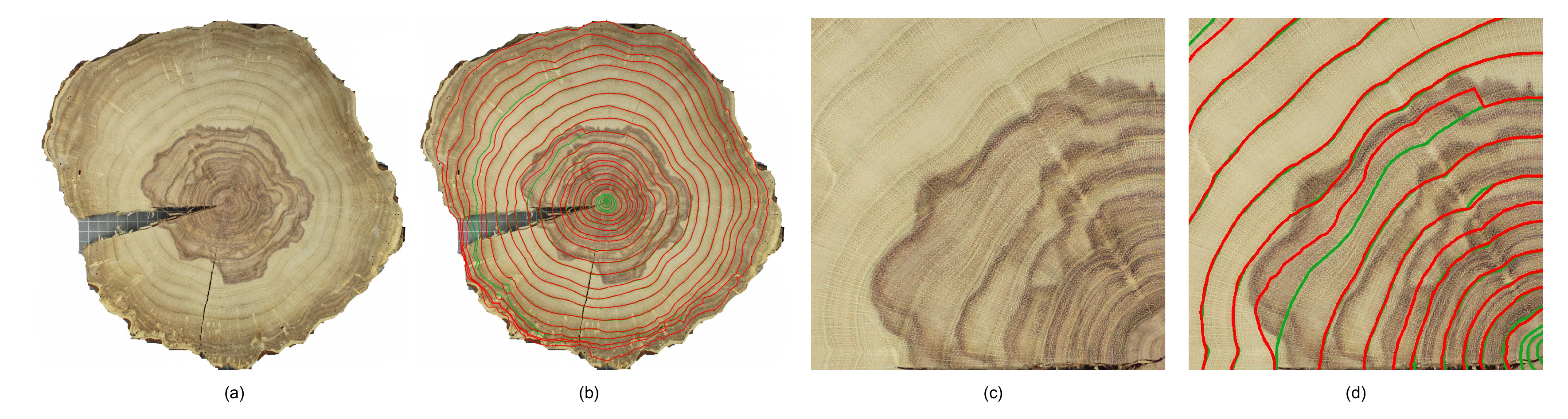}
    \caption{G. triachantos image sample G6 from \cite{deepcstrd}. (a) Sample after background removal. (b) Expert-delineated rings are shown in green, and automatically detected rings are shown in red. Correct detections overlap with the expert annotations. (c, d) Zoomed-in views highlighting a disk's region.}
    \label{fig:g6}
\end{figure}

\begin{figure}
    \centering
    \includegraphics[width=\linewidth]{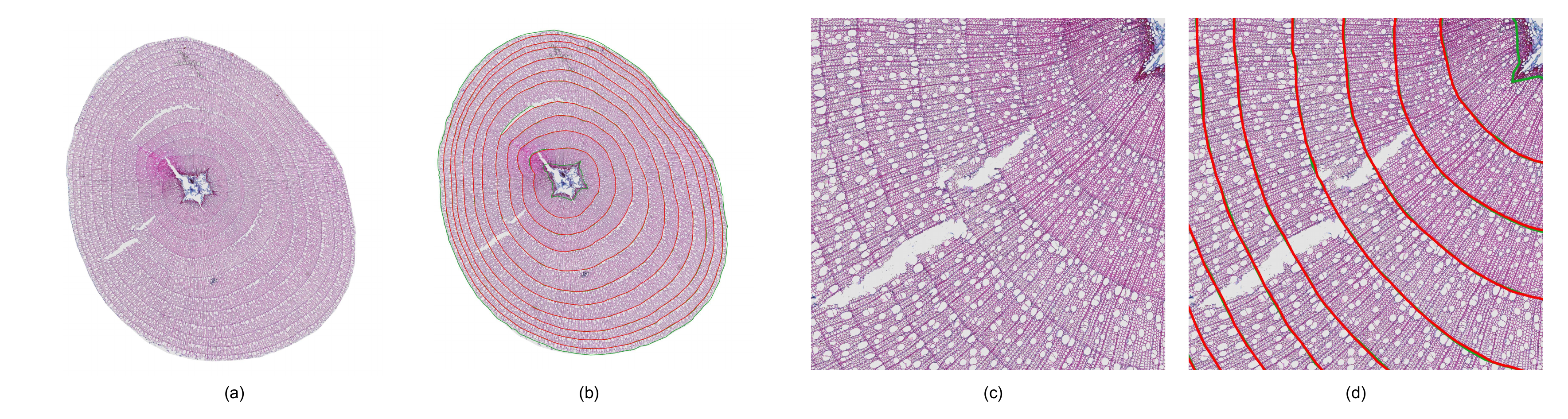}
    \caption{S. glauca image sample 1W23S20 from \cite{salix}. (a) Sample after background removal. (b) Expert-delineated rings are shown in green, and automatically detected rings are shown in red. Correct detections overlap with the expert annotations. (c, d) Zoomed-in views highlighting a disk's region.}
    \label{fig:salix}
\end{figure}

\begin{figure}
    \centering
    \includegraphics[width=\linewidth]{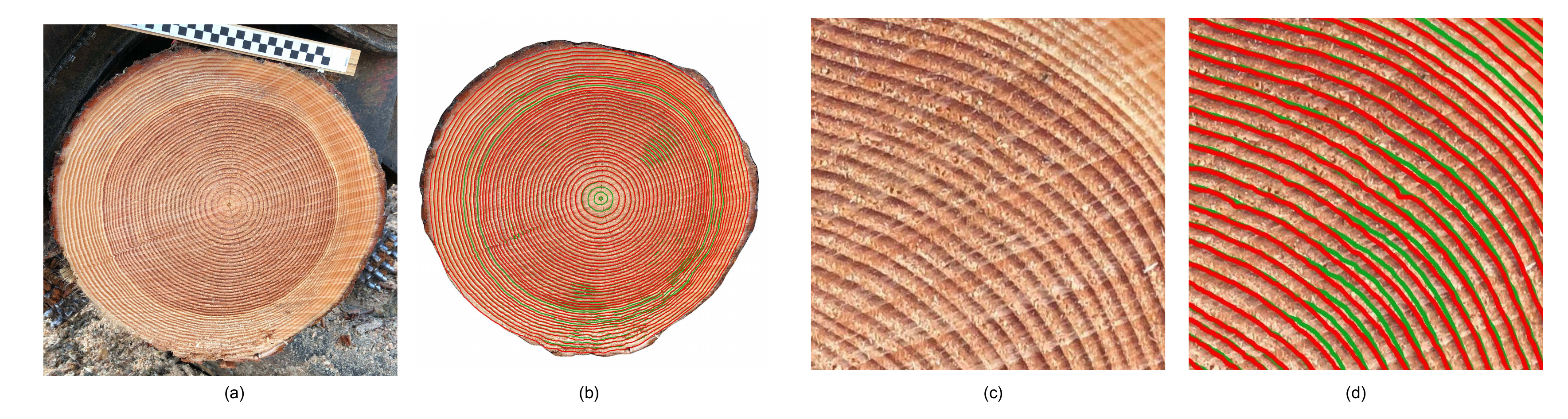}
    \caption{Douglas fir image sample C07c from \cite{DouglasFir}. (a) Image sample. (b) Expert-delineated rings are shown in green, and automatically detected rings are shown in red. Correct detections overlap with the expert annotations. (c, d) Zoomed-in views highlighting a disk's region.}
    \label{fig:douglas}
\end{figure}

The application of TRAS to external datasets (\Cref{fig:c14} to \Cref{fig:douglas}) demonstrates its versatility across different species, imaging conditions, and wood surface qualities. The successful detection of rings in \textit{G.\ triacanthos} (\Cref{fig:g6}), \textit{S.\ glauca} (\Cref{fig:salix}), and Douglas fir (\Cref{fig:douglas}) samples indicates that the integrated algorithms, particularly the deep learning-based methods trained on \textit{P.\ taeda}, exhibit reasonable generalization to other coniferous and broadleaf species.


The samples from \citep{CSTRD} (\Cref{fig:f03d} and \Cref{fig:l02b}) present additional challenges, including cracks and irregular growth patterns, yet TRAS's automatic methods produced detections that closely matched expert annotations. These results suggest that the combination of multiple detection algorithms increases the likelihood that at least one method will perform well on a given sample, providing users with flexibility in selecting the most appropriate approach for their specific dataset.

However, it is important to acknowledge that performance may vary when applied to species with significantly different anatomical characteristics, such as ring-porous hardwoods or tropical species with indistinct ring boundaries. Future work should evaluate TRAS on a broader taxonomic range to establish its limitations and guide the development of species-specific models.

\subsection{Advantages of Area-Based Measurements}

A key contribution of TRAS is its capability to compute area-based metrics, which are particularly relevant for samples exhibiting high pith eccentricity. In such cases, ring width measurements along a single radial path may not accurately represent annual growth, as demonstrated by \citep{atmos14020319}. The equivalent ring width metric implemented in TRAS (computed from ring area) provides a more robust indicator of volumetric growth that accounts for asymmetric ring formation.

The ability to compute cumulative earlywood and latewood areas (\Cref{fig:cumulative}) enables researchers to investigate the differential response of these wood components to environmental variables. Studies by \citep{latewood_1,latewood_2} have shown that latewood width often exhibits stronger correlations with climate variables than total ring width, supporting the utility of TRAS's separate measurement of these components. Furthermore, the Circle Similarity Factor and Ring Eccentricity metrics provide quantitative assessments of growth symmetry, which can reveal long-term trends in competitive asymmetry or responses to directional environmental stresses.

\subsection{Efficiency Gains and User Experience}

The semi-automatic workflow implemented in TRAS significantly reduces the time required for tree ring analysis compared to fully manual approaches. Although precise timing comparisons were not conducted in this study, the high automatic detection rates (F-scores around 80\%) indicate that users need only correct approximately 20\% of detections, representing a substantial time saving relative to manual delineation of all rings. The integrated LabelMe-based annotation canvas provides an intuitive environment for these corrections, allowing users to add, delete, or adjust rings with minimal training.

The modular design of TRAS, with separate menus for image preprocessing, ring detection, editing, and metrics computation, facilitates a logical workflow while maintaining flexibility. Users can, for instance, skip automatic detection entirely and manually delineate rings if preferred, or they can test multiple detection algorithms sequentially to identify the best-performing method for their samples.

\subsection{Limitations and Future Improvements}

Despite its strengths, TRAS has several limitations that should be addressed in future development. The current installation process requires users to interact with the command line, which may present a barrier for researchers without programming experience. Developing a standalone executable application with a graphical installer would significantly improve accessibility and adoption. The deep learning models integrated into TRAS were trained primarily on coniferous species, particularly \textit{P.\ taeda}; expanding training datasets to include diverse taxonomic groups and imaging conditions would enhance the generalization capabilities of DeepCS-TRD and INBD.

For users wishing to improve detection performance on their specific datasets, TRAS supports uploading custom-trained DeepCS-TRD and INBD models. Step-by-step documentation for fine-tuning these models using TRAS-corrected annotations is provided in the \href{https://github.com/hmarichal93/tras/blob/main/docs/training_custom_models.md}{project repository}.

TRAS currently provides limited support for analyzing time-series data or cross-dating samples. Integrating statistical tools for de-trending, cross-correlation, and chronology development would position TRAS as a complete dendrochronological analysis platform. Implementing export functionality to common dendrochronological data formats (e.g., Tucson format) would further facilitate integration with existing analysis pipelines. Future versions should also explore the integration of machine learning models capable of detecting and classifying wood defects (e.g., compression wood, reaction wood, false rings) automatically.

\subsection{Implications for Dendrochronological Research}

TRAS addresses a critical need in modern dendrochronology for efficient, reproducible, and comprehensive analysis of tree ring patterns. By providing open-source code and a user-friendly interface, TRAS facilitates collaboration, method validation, and data sharing within the research community.

The area-based metrics computed by TRAS enable investigations into growth dynamics that were previously impractical with traditional tools focused on linear measurements. Researchers studying trees with irregular growth patterns, reconstructing historical climate conditions from eccentrically growing specimens, or investigating competitive interactions in dense forest stands will particularly benefit from these capabilities. Moreover, the ability to exclude regions affected by wood defects from area calculations ensures that growth estimates reflect true biological responses rather than being confounded by structural anomalies.

In conclusion, TRAS represents a significant advancement in dendrochronological methodology, combining state-of-the-art automatic detection algorithms with flexible manual correction tools and comprehensive measurement capabilities. The validation results demonstrate that TRAS produces accurate measurements comparable to established tools while offering substantial improvements in efficiency and analytical depth. With continued development and community engagement, TRAS has the potential to become a standard tool for tree ring analysis across diverse ecological and climatological applications.

\section{Conclusion}


We presented TRAS, an open-source graphical user interface tool for automatic tree-ring delineation. It enables users to correct automatic detections and manually compute accurate ring-based measurements. The tool integrates three state-of-the-art automatic detection algorithms (CS-TRD, DeepCS-TRD, and INBD) implemented in Python.



TRAS facilitates the analysis of ring growth in samples where eccentricity limits the applicability of core-based measurements, providing a more comprehensive and efficient alternative. To our knowledge, it is the first graphical tool to combine two-dimensional automatic tree-ring detection with manual correction capabilities on cross-sectional images.

 TRAS is under active development. The current version (v2) is available as a cross-platform desktop application for Windows, macOS, and Linux, and includes a command-line interface (\texttt{tras\_detect}) for batch processing. Future work will focus on extending support to additional tree species and image acquisition setups, integrating more recent deep learning architectures, and further reducing the manual effort required for ring correction.

\subsection{Funding}
This work was supported by project ANII-FMV-176061: UruDendro 2.0: Aplicación de técnicas de procesamiento de imágenes e inteligencia artificial para la dendrometría automática de especies de madera nativas y comerciales. 

\subsection{Acknowledgement}

The experiments presented in this paper were conducted using ClusterUY. We also gratefully acknowledge the collaboration of Candice C. Power, Verónica Casaravilla, Ludmila Profumo and Christine Lucas.

\subsection{Conflicts of Interest}
The authors declare that they have no conflict of interest.

\subsection{Data Availability Statement}

All data and materials used in this study are openly available in the TRAS public repository at \url{https://hmarichal93.github.io/tras}. The repository contains the software source code, image datasets, and supporting scripts required to reproduce the analyses and figures presented in this manuscript.

\appendix

\section{Keyboard Shortcuts}
\label{sec:shortcuts}

Table~\ref{tab:shortcuts} lists all keyboard shortcuts available in TRAS. Shortcuts can be customized by editing the user configuration file \texttt{\textasciitilde/.trasrc}.

\begin{table}[h]
\centering
\small
\caption{Keyboard shortcuts available in TRAS.}
\label{tab:shortcuts}
\begin{tabular}{ll}
\toprule
\textbf{Action} & \textbf{Shortcut} \\
\midrule
\multicolumn{2}{l}{\textit{File Operations}} \\
Open image / label file    & Ctrl+O \\
Open directory             & Ctrl+U \\
Next image                 & D \textit{or} Ctrl+Shift+D \\
Previous image             & A \textit{or} Ctrl+Shift+A \\
Save                       & Ctrl+S \\
Save as                    & Ctrl+Shift+S \\
Close file                 & Ctrl+W \\
Delete file                & Ctrl+Delete \\
Quit application           & Ctrl+Q \\
\midrule
\multicolumn{2}{l}{\textit{TRAS Workflow}} \\
Sample metadata            & Ctrl+M \\
Set image scale            & Ctrl+L \\
Preprocess image           & Ctrl+I \\
Detect tree rings          & Ctrl+T \\
Measure ring width         & Ctrl+G \\
View ring properties       & Ctrl+K \\
\midrule
\multicolumn{2}{l}{\textit{Drawing \& Editing}} \\
Draw closed ring (polygon) & Ctrl+N \\
Draw open ring (linestrip) & Ctrl+Shift+N \\
Edit mode                  & Ctrl+J \\
Edit label                 & Ctrl+E \\
Delete ring                & Delete \\
Duplicate ring             & Ctrl+D \\
Copy ring                  & Ctrl+C \\
Paste ring                 & Ctrl+V \\
Undo                       & Ctrl+Z \\
Remove selected point      & Backspace \textit{or} Meta+H \\
Add point to existing ring & Alt+Click (on edge, in Edit mode) \\
\midrule
\multicolumn{2}{l}{\textit{View \& Zoom}} \\
Zoom in                    & Ctrl++ \textit{or} Ctrl+= \\
Zoom out                   & Ctrl+- \\
Zoom to 100\%              & Ctrl+0 \\
Fit to window              & Ctrl+F \\
Fit width                  & Ctrl+Shift+F \\
Toggle all rings visibility & T \\
\midrule
\multicolumn{2}{l}{\textit{Help}} \\
Show shortcuts dialog      & F1 \\
\bottomrule
\end{tabular}
\end{table}

\end{document}